\documentclass[sigconf]{acmart}
\usepackage{amsmath}
\usepackage{algorithm}
\usepackage[noend]{algpseudocode} 
\usepackage{mathtools}
\usepackage{subfigure}
\usepackage{leftidx}
\usepackage{booktabs} 
\usepackage{todonotes}
\usepackage{bbding}
\usepackage{multirow}
\usepackage{threeparttable}
\usepackage{listings}

\def\BibTeX{{\rm B\kern-.05em{\sc i\kern-.025em b}\kern-.08emT\kern-.1667em\lower.7ex\hbox{E}\kern-.125emX}}

\renewcommand\footnotetextcopyrightpermission[1]{} 
\begin{document}

%
\title{FamilySeer: Towards Optimized Tensor Codes by Exploiting Computation Subgraph Similarity}

%
\author{Shanjun Zhang, Mingzhen Li, Hailong Yang, Yi Liu, Zhongzhi Luan and Depei Qian}
\affiliation{
  \institution{School of Computer Science and Engineering\\
  Beihang University, Beijing, China, 100191}
}
\email{{lu_cheung, lmzhhh, hailong.yang, yi.liu, 07680, depeiq}@buaa.edu.cn}
%

%

\begin{abstract}

Deploying various deep learning (DL) models efficiently has boosted the research on DL compilers.
The difficulty of generating optimized tensor codes drives DL compiler to ask for the auto-tuning approaches, and the increasing demands require increasing auto-tuning efficiency and quality.
Currently, the DL compilers partition the input DL models into several subgraphs and leverage the auto-tuning to find the optimal tensor codes of these subgraphs. 
However, existing auto-tuning approaches usually regard subgraphs as individual ones and overlook the similarities across them, and thus fail to exploit better tensor codes under limited time budgets.

We propose FamilySeer, an auto-tuning framework for DL compilers that can generate better tensor codes even with limited time budgets. 
FamilySeer exploits the similarities and differences among subgraphs can organize them into subgraph families, where the tuning of one subgraph can also improve other subgraphs within the same family.
The cost model of each family gets more purified training samples generated by the family and becomes more accurate so that the costly measurements on real hardware can be replaced with the lightweight estimation through cost model. 
Our experiments show that FamilySeer can generate model codes with the same code performance more efficiently than state-of-the-art auto-tuning frameworks. 

\end{abstract}


%
%

%

%

%
\maketitle

\renewcommand{\algorithmicrequire}{\textbf{Input:}}
\renewcommand{\algorithmicensure}{\textbf{Output:}}

\newcommand{\tabincell}[2]{\begin{tabular}[t]{@{}#1@{}}#2\end{tabular}}

\lstset{
	basicstyle          =   \footnotesize\sffamily,          
	keywordstyle        =   \bfseries,          
	commentstyle        =   \rmfamily\itshape,  
	stringstyle         =   \ttfamily,  
	numbers             =   left,   
	showspaces          =   false,  
	numberstyle         =   \ttfamily,    
	showstringspaces    =   false,
	captionpos          =   t,      
	frame               =   lrtb,   
	breaklines, 
	columns=flexible, 
}





\newcommand{\red}[1] {\textcolor{red}{\it{#1}}}
\newcommand{\revision}[1] {\textcolor{blue}{{#1}}}
\newcommand{\secondrevision}[1] {\textcolor{black}{{#1}}}
\newcommand{\shepherd}[1] {\textcolor{black}{{#1}}}
\newcommand{\finalshepherd}[1] {\textcolor{black}{{#1}}}
\newcommand{\remove}[1] {{\textcolor{red}{\sout{#1}}}}
\newcommand{\add}[1] {{\textcolor{red}{\underline{#1}}}}
\newcommand{\egg}[1] {}
\newcommand{\separate}[1] {\textbf{\center ======  #1  ====== }}
\newcommand{\smalltitle}[1] {\vspace{6pt} \noindent \textbf{#1}}
\def\approx{$\sim$}

\def\receivers{\textbf Q}
\def\capacity{{ \textbf{c}}}
\def\upperBound{{\mathcal D}}
\def\schedule{{\mathcal S}}
\def\throughput{{\cal T}}
\def\period{{\textbf Z}}

\newcommand{\prob}[1]{{\textbf{Pr}\left(#1\right)}}
\newcommand{\mcell}[2]{ \parbox[h]{#1}{ \vspace{0.5mm} #2 \vspace{0.5mm}}}



\renewcommand{\algorithmicrequire}{\textbf{Input:}}
\renewcommand{\algorithmicensure}{\textbf{Output:}}

\newcommand{\KB}{~KB}
\newcommand{\MB}{~MB}
\newcommand{\GB}{~GB}
\newcommand{\MBs}{~MB/s}
\newcommand{\mus}{~$\mu s$}
\newcommand{\ms}{~$ms$}

\newcommand{\eg}{\textit{e.g.}}
\newcommand{\ie}{\textit{i.e.}}
\newcommand{\etc}{\textit{etc.}}
\newcommand{\aka}{\textit{a.k.a.}}

\section{Introduction}
\label{sec:introduction}

The performance of deep learning (DL) models is critical to address the ever-increasing demands in the fields of computer vision~\cite{wang2014face}, natural language processing~\cite{singh2017machine}, auto-driving~\cite{kore2019obstacle} and etc. However, manually optimizing the deep learning models is both error-prone and hardly-portable due to the model diversity and hardware complexity. Therefore, deep learning compilers play an important role to enable high-performant code generation for various models on different hardware automatically. Currently, several deep learning compilers have been proposed such as XLA~\cite{abadi2017computational}, nGraph~\cite{cyphers2018intel}, Tensor Comprehension~\cite{vasilache2018tensor}, TVM~\cite{chen2018tvm} and etc. A compiler takes the DL models from DL frameworks (e.g., Tensorflow~\cite{abadi2016tensorflow}, Mxnet~\cite{chen2015mxnet}, Pytorch~\cite{paszke2019pytorch}) as input.
It converts the model into multiple level of intermediate representations (IRs), and then automatically applies various performance optimizations regarding the model characteristics and underlying hardware in order to generate high-performant model codes~\cite{dlcompiler}. Although different design philosophies have been adopted in different compilers, the fundamental procedures to generate efficient model codes are similar. These procedures can be commonly divided into two phases including: \textit{1)} integrating a collection of code transformations (optimization techniques), and \textit{2)} determining the optimal sequence to apply the code transformations (searching techniques).

During compilation, the deep learning model is first transformed into a computation graph, where each node represents an operator (e.g., convolution and ReLU) and each edge represents the data flow. The computation graph is further divided into subgraphs, where each subgraph may contain several operators that can be fused together (e.g., convolution, ReLU and pooling). The code generation procedures are then applied to each subgraph to determine the optimal sequence of code transformations for high-performant codes. The optimization techniques have commonly been adopted including loop tiling, loop fusion, parallelization, vectorization and etc. Whereas the searching techniques have commonly been adopted including AutoTVM~\cite{chen2018learning}, Ansor~\cite{zheng2020ansor}, MetaTune~\cite{ryu2021metatune} and etc. The fundamental idea of searching techniques can be commonly divided into three step: \textit{1)} explore massive code transformations and generate huge amount of code candidates, which forms a large search space, \textit{2)} apply searching algorithms to identify the candidates with high performance potential, and \textit{3)} evaluate these candidates on real hardware and fine-tune the searching algorithms to identify better code candidates. The above three steps are applied iteratively until the time budget is expired. To maintain the search overhead within acceptable time budget, the searching techniques usually allocate the limited time slots proportionally to the subgraphs based on their dominance in the model execution time.

With the optimization techniques well studied in existing deep learning compilers, the searching techniques become critical in generating high-performant model codes. A good searching technique needs to generate large enough search space (affecting search quality) and explore the search space efficiently (affecting search speed). For example, the recently proposed searching scheme Ansor~\cite{zheng2020ansor} adopts code sketch and parameter sampling to ensure good search quality through a large search space. In addition, it adopts a cost model to quickly filter out code transformation sequence with high performance potential, and thus accelerates the search process. Since the accuracy of cost model can directly affect the search efficiency, several works~\cite{steiner2021value,baghdadi2021deep,kaufman2021learned} have been proposed to improve the accuracy of cost model through either careful feature engineering or relying on machine learning methods. However, the above works universally adopt a single cost model for all subgraphs, that fails to achieve accurate performance estimation for subgraphs with diverse characteristics, and further constrains the search quality to generate better codes. In addition, existing searching techniques attempt to allocate more search time to the bottleneck subgraphs, and thus the cost models are trained with bias towards these subgraphs. Such search policy neglects the optimization potential of other subgraphs and slows down the convergence to optimal results for all subgraphs of the entire model.

To address the above limitations, we propose \textit{FamilySeer}, a new search method to determine the optimal sequence of code transformations during compilation. The fundamental idea of \textit{FamilySeer} is to exploit the similarity of subgraphs, and organize the similar subgraphs into a collection of subgraph families. With the subgraph families identified, \textit{FamilySeer} constructs cost models at subgraph family basis, which can effectively address the diverse characteristics across different subgraph families and thus improve the cost model accuracy. In addition, the subgraphs within a family can share the search result during each tuning iteration without applying the costly measurements on real hardware, which can speedup the search process to converge to optimal results within limited time budget. We implement \textit{FamilySeer} within the deep learning compiler TVM to enable model code generation across different hardware platforms. Moreover, we implement auxiliary optimizations such as parallelization of cost model training and code measurement on GPUs, which can further speedup the search process.

Specifically, this paper makes the following contributions:
\begin{itemize}
	\item We propose \textit{FamilySeer}, a new auto-tuning framework that can be applied during model compilation to generate more efficient model codes. \textit{FamilySeer} exploits the subgraph similarity to form a collection of subgraph families, and construct cost models at subgraph family basis to improve cost model accuracy.
	\item We re-design the search process to better utilize the advantage of subgraph families. Particularly, we enable the subgraphs within each family to share the search results within each tuning iteration, avoiding costly code measurements on real hardware and thus accelerating the search process to converge to optimal results.
	\item We evaluate the effectiveness of \textit{FamilySeer} with representative models on both CPU and GPU platforms. The experimental results demonstrate that, compared to the state-of-the-art search scheme Ansor, \textit{FamilySeer} can generate model codes more efficient with {2.49$\times$ and 3.04$\times$} performance speedup on average within the same code performance.
\end{itemize}

The rest of the paper is organized as follows. Section~\ref{sec:background} describes the background of search-based auto-tuning framework in deep learning compiler. Section~\ref{sec:motivation} presents the drawback of auto-tuning ignored by the deep learning compiler. Section~\ref{sec:overview} and Section~\ref{sec:methodology} present the design overview and our detailed implementation of our \textit{FamilySeer}. Section~\ref{sec:evaluation} presents the evaluation results and compares the search quality and efficiency with the state-of-the-art search scheme. Section~\ref{sec:conclusion} concludes this paper.

\section{Background}
\label{sec:background}

The searching techniques of DL compilers are critical to generating high-performance model codes. TVM is the state-of-the-art deep learning compiler that applies many search-based optimization techniques. Ansor is the second generation searching technique of TVM. This work is built on top of the Ansor. To better understand the searching procedure, we take Ansor as an example for illustration.

\textbf{Generating Subgraphs - } Ansor is implemented within the deep learning compiler TVM. TVM compiles a DL Model into a computation graph, where a node represents an operator, and the edges represent the data flow. TVM then fuses these operators according to the pre-defined optimization rules.
Ansor takes these fused operators and regards them as a subgraph. As an operator can take tensors with various shapes as the input (e.g., in ResNet50\_v1, the input shape of a convolution operator varies from $7\times7\times512$ to $56\times56\times64$), many subgraphs may have the same operator sequence but with different input shapes. We have summarized the number of subgraphs the state-of-the-art DL models in Table ~\ref{tab:SubgraphNumberType}.

\begin{table}
\centering
\caption{The number of the subgraphs partitioned by TVM of state-of-the-art DL models.}
\label{tab:SubgraphNumberType}
\begin{tabular}{cc}
    \hline
    DL Models          & Number of Subgraph  \\ \hline
    ResNetv1          & 25$\sim$28          \\
    ResNetv2          & 30$\sim$32          \\
    Mobilenet          & 22$\sim$25          \\
    Mobilenetv2        & 34$\sim$38          \\
    BERT               & 11$\sim$13          \\
    RoBERTa            & 9$\sim$11           \\
    GPT2               & 10                  \\
    Vision Transformer                & 13                  \\ \hline
    \end{tabular}
\end{table}

\textbf{Scheduling Subgraphs - } Ansor adopts the gradient descent algorithm to schedule the evaluation process of subgraphs, which allocates the limited time slots to the subgraphs, in order to improve the search efficiency and the search quality.
For example, GPT2 is divided into 10 subgraphs. But only three subgraphs contribute to 80\% of the overall execution time. Thus Ansor allocates more time to these subgraphs than others.
Specifically, Ansor passes each subgraph's performance information (execution time, GFlops, \etc) into the gradient descent algorithm and allocates the next time slot to the subgraph with lowest negative gradient.
This algorithm behaves well in the first few iterations since the bottleneck subgraphs have more possibility to acquire the time slots.
But as the bottleneck subgraph reaches its performance ceiling, their potential of further performance improvement is minor. However, they are still allocated with more time slots according to this algorithm, which hinders the evaluation process of other subgraphs with more potential.

\textbf{Building the Cost Model - }
To reduce the overhead of evaluating the massive transformed codes for each subgraph on real hardware, cost models have been adopted to estimate the performance of the transformed codes. Ansor adopts an XGBoost cost model to score each subgraph, which has been designed with heavy feature engineering to correctly predict the performance of transformed codes. The subgraphs with top K scores are selected to evaluate on real hardware. These subgraph measurements are accumulated and then used to train a new cost model. However, at the beginning of code search, when there is no enough training data, the cost model is hardly accurate to identify high-performant code candidates. In addition, compounded with the subgraph scheduling strategy that favors bottleneck subgraphs, the cost model is gradually trained with biased data, which eventually constrains the model accuracy.

\textbf{Evaluating Code Candidates - } The potential code candidates (transformed codes) of each subgraph selected by the cost model are then evaluated on real hardware. The real measurements are used to train the cost model as well as adjust the subgraph scheduling decisions. One limitation of current Ansor implementation is that it evaluates the code candidates sequentially, and fails to exploit the independence among the code candidates for parallel evaluation. Since the multi-GPU platforms become ubiquitous in deep learning applications, evaluating the independent code candidates simultaneously on multiple GPUs can significantly boost the search process.

\section{Motivation}
\label{sec:motivation}

In this section, we present the observations of inefficiencies from the existing auto-tuning frameworks for DL models, and we analyze the drawbacks of these frameworks to motivate the design of FamilySeer.

\subsection{Overlooking Similarities and Differences Across Subgraphs}

\begin{figure}[htbp]
	\centering
	\includegraphics[width=0.95\linewidth]{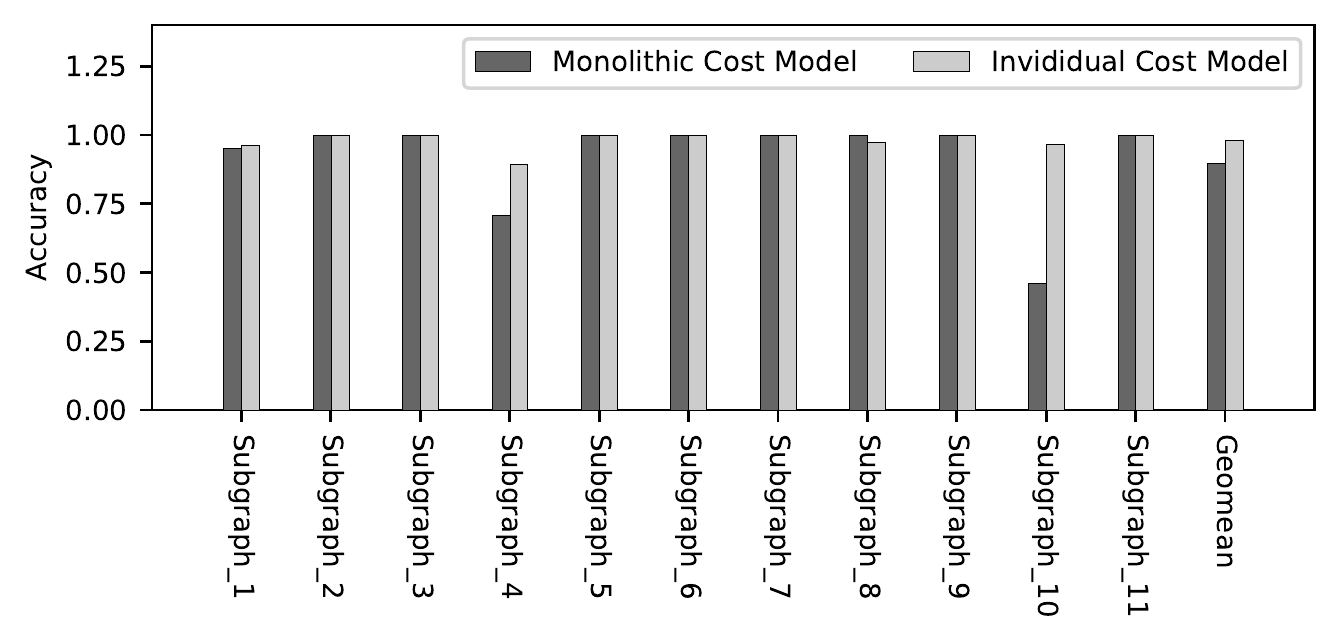}
	\caption{Accuracy difference between using monolithic cost model and individual cost models.}
	\label{fig:acc_bar}
\end{figure}

The auto-tuning frameworks usually adopt the cost model to estimate the performance of the transformed programs. Specifically, they build a monolithic cost model to estimate all transformed programs of all subgraphs during the auto-tuning~\cite{zheng2020ansor, zheng2021tenset}.
However, due to the variety of subgraphs,  the cost model may fail to preform accurate estimations across all programs and introduce great bias between estimated performance and real performance, which can reduce the efficiency of the auto-tuning procedure.
Figure ~\ref{fig:acc_bar} shows the experiment of BERT-Large about the cost model accuracy. The BERT-Large can be partitioned into 11 subgraphs, and we select 256 samples (candidates) for each subgraph to build the monolithic cost model ($256\times 11$ samples in total).
It is obvious that the prediction accuracy (black bars) on subgraph\_4 and subgraph\_10 is much lower than that of other nine subgraphs. 

Since the cost model is trained from scratch, its accuracy is depending on the quality of the training samples significantly. We suspect that the training samples from different subgraphs can not only cooperate with each other, which leads to improved accuracy on some subgraphs but also interfere with each other, which leads to decreased accuracy.
To better understand this phenomenon, we further equip each subgraph with a individual cost model, and train the model with the subgraph's samples. That is, each individual model is trained with purified samples (\ie, $\frac{1}{11}$ of the original samples).
As shown in Figure ~\ref{fig:acc_bar} (gray bars), the accuracy on subgraph\_4 and subgraph\_10 returns to normal, which is similar with other subgraphs.
This is because the cost models are trained with purified samples without any biased sample. 
However,  the accuracy on subgraph\_8 decreases because its cost model cannot leverage the samples from similar subgraphs and thus suffers from decreased samples.

\begin{figure}[htbp]
	\centering
	\includegraphics[width=0.9\linewidth]{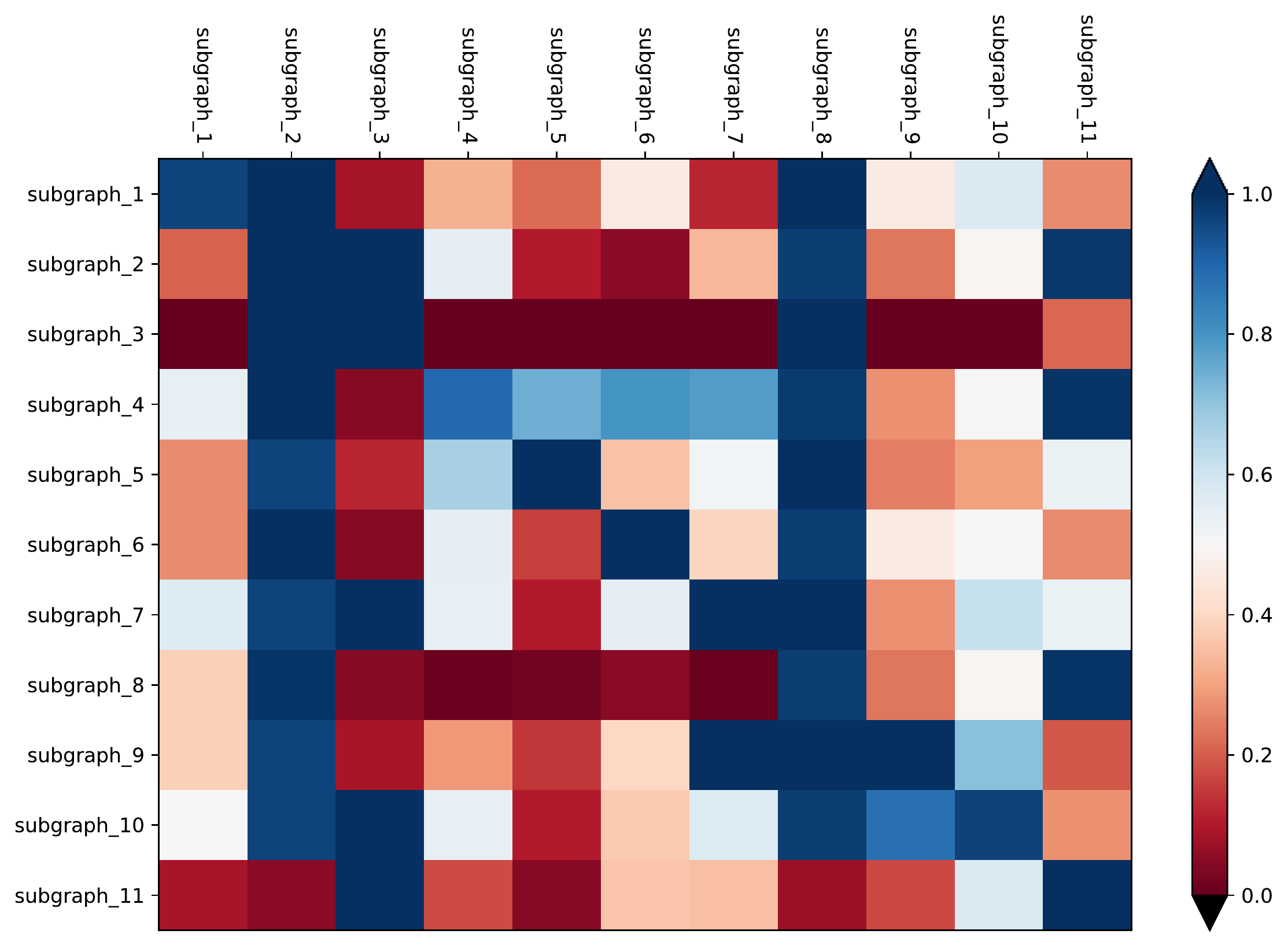}
	\caption{Accuracy heatmap among difference subgraphs, where the cell in row $X$ and column $Y$ represents the accuracy when using the individual cost model of subgraph\_X to predict the validation samples of subgraph\_Y.}
	\label{fig:heatmap}
\end{figure}

Moreover, we conduct an experiment to explore the relationship across training samples from different subgraphs.  
As shown in Figure~\ref{fig:heatmap}, the cell in row $X$ and column $Y$ represents the accuracy when using the individual cost model of subgraph\_X to predict the validation samples of subgraph\_Y.  
It is obvious that the relationship across the subgraphs is complicated enough. For example, the cost model of subgraph\_4 can predict the samples of subgraph\_2/4/5/6/7/8/11 with an accuracy greater than 74.5\%. However, only the cost model of subgraph\_6 can predict the samples subgraph\_2
with an accuracy greater than 99.8\%. 
Besides, the samples of subgraph\_2/8 can be accurately predicted by cost models of other subgraphs (expect subgraph\_11).  
Therefore, the similarities and differences between subgraphs are complex enough, and the currently adopted monolithic cost models overlook these features, missing the possibility of extra performance improvement.

\subsection{Wasting Time on Subgraphs Without Potential}

The auto-tuning frameworks allocate a predefined time budget for the tuning procedure. They consume the time budget to tune the subgraphs in an iterative manner, till using up the time budget.
In each iteration, they tune single subgraph. They generate multiple program candidates for each subgraph, and then sort them in ascending order, according to their latency estimated by the cost model. They select the top-k program candidates (\eg, top-64 in Ansor), and measure their latency in real hardware (\eg, V100 GPU), which is called \textit{measurements}. Notably, each measurement consumes a time budget. 
The results of the \textit{measurements} are used to train the cost models, at the same time, the results are normalized to represent the real latency of the subgraph.
These frameworks adopt the greedy algorithm to allocate time budgets to the subgraphs with higher latency (\ie, bottleneck subgraphs).
However, if a bottleneck subgraph has trivial headroom of latency improvement, the greedy algorithm still prioritizes this subgraph, which wastes the budget and leaves other subgraphs far from the optimal. 

\begin{figure}[htbp]
	\centering
	\includegraphics[width=0.85\linewidth]{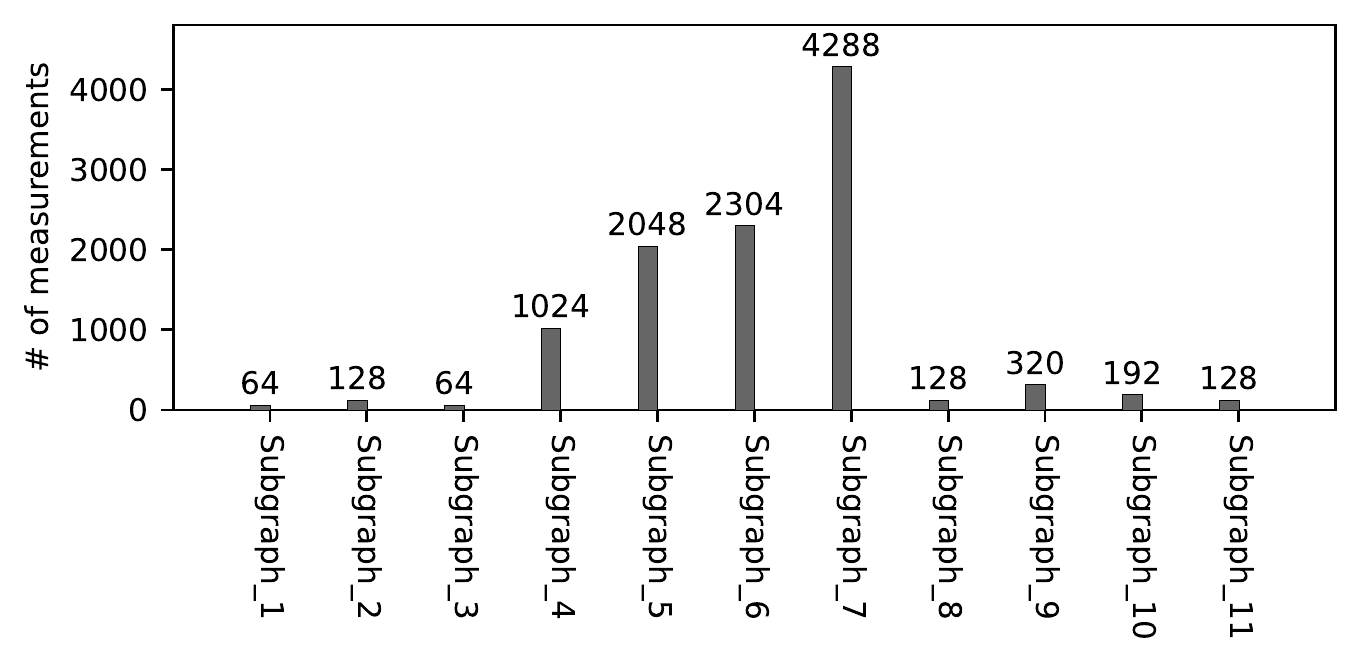}
	\caption{Time budget allocation among different subgraphs.}
	\label{fig:motivation_trials_0}
\end{figure}

\begin{figure}[htbp]
	\centering
	\includegraphics[width=0.8\linewidth]{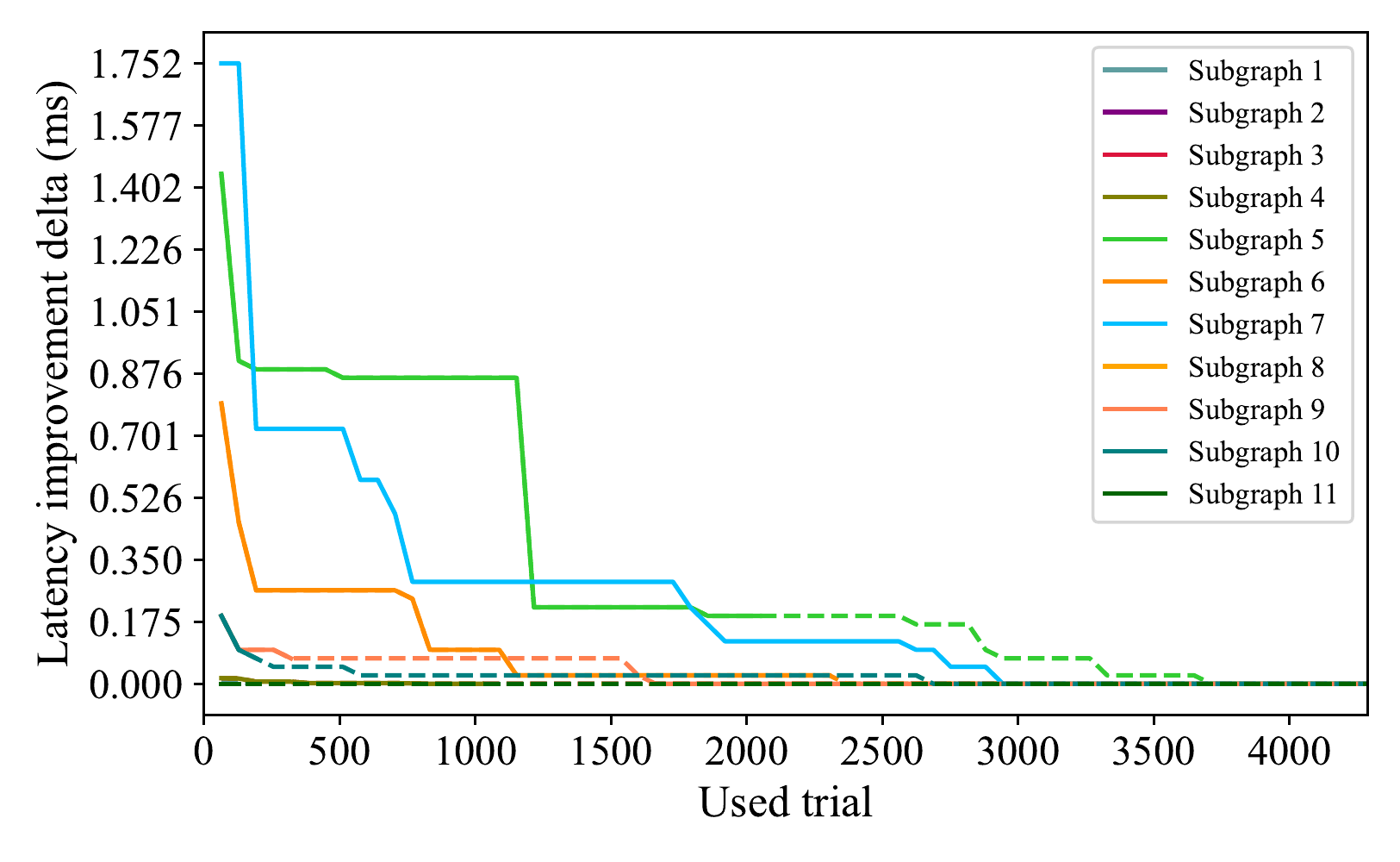}
	\caption{The latency improvement of each subgraph when allocating sufficient time budgets.}
	\label{fig:motivation_trials_1}
\end{figure}

We tune the BERT-Large model with Ansor on a V100 GPU, where the time budget is set to 9900 measurements as Ansor recommended. 
As shown in Figure ~\ref{fig:motivation_trials_0}, subgraph\_7 gets the most time budget, 4288 measurements, and subgraph\_4/5/6 gets 1024, 2048, 2304 measurements, respectively. While other subgraphs get much fewer measurements.
As shown in Figure ~\ref{fig:motivation_trials_1}, we further allocate extra time budgets for these subgraphs, so that all subgraphs have the same budget (\ie, 4288 measurements) as the subgraph\_7. The dashed lines indicate the latency improvement with the extra budgets. Notably, subgraph\_2/8 have limited number of program candidates (320 and 320, respectively) in total and all the candidates have already been measured. We have several observations:
\textit{1)} Subgraph\_7 has no improvement during the 2944-4288 measurements.
\textit{2)} Subgraph\_5 has the highest improvement for 0.192 ms.
\textit{3)} If we just reallocate the wasted budgets to other subgraphs evenly, the overall latency of the subgraphs can improve 0.049 ms.

To summarize,  we believe that the existing auto-tuning frameworks fail to utilize both the training samples and the time budgets and leads to the unsatisfying tuning efficiency, which motivates us to design an efficient auto-tuning framework.

\section{Design Overview}
\label{sec:overview}
\begin{figure}[htbp]
	\centering
	\includegraphics[width=\linewidth]{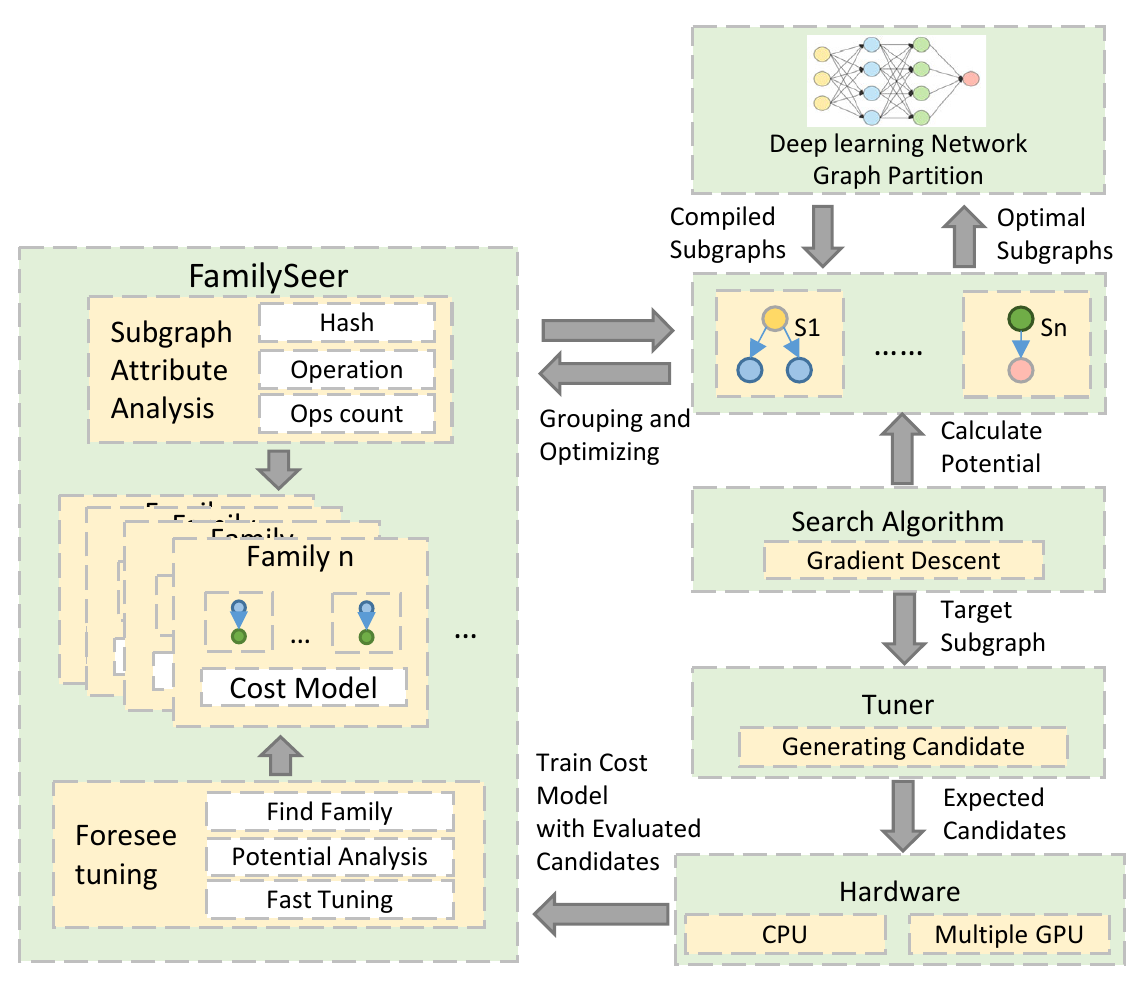}
	\caption{The overall design of FamilySeer, including the subgraph family and the foresee tuning.}
	\label{fig:overview}
\end{figure}

As stated above, the training samples and the time budgets are underutilized by existing auto-tuning frameworks of deep learning compilers. 
We believe that the auto-tuning frameworks can achieve better efficiency (\ie, less searching time) and better quality (\ie, more inference throughput) 
even with the same amount of samples and budgets. 
Therefore, we propose a new auto-tuning framework, FamilySeer, which reschedules the training samples and time budgets for superior performance to other frameworks. 
FamilySeer takes the deep learning model as the input and obtains a series of subgraphs according to the graph partition rules (\eg, rules provided by TVM). It clusters the subgraphs into several families, where the subgraphs in a family can share the training samples and time budgets. 
Then it tunes the subgraph families by generating program candidates and select the candidates with less latency, in an iterative manner. 
Notably, FamilySeer focuses on the improvement of subgraph family rather than that of individual subgraphs.
Specifically, the design of FamilySeer primarily contains two parts: the subgraph family and the family performance tuner (\ie, foresee tuning), as shown in Figure~\ref{fig:overview}.

The subgraph family is the foundation for the improvement of FamilySeer's tuning efficiency. 
Forming the subgraph families needs to cluster the subgraphs into the subgraph families, and construct of the cost model to estimate the latency of generated programs candidates within the family. 
In order to avoid introducing extra overhead to the auto-tuning process  we analyze the similarity of subgraphs according to their attributes and form the families accordingly. 
During the auto-tuning, the search algorithm finds the subgraph family to which the current tuning subgraph belongs, and then filter the various program candidates to find the candidates with more improving potential with the help of the family's cost model.

The family performance tuner optimizes the tuning process of the subgraph family. After the tuner figures out the bottleneck subgraphs with higher latency, it generates program candidates of the subgraphs based on the time budget. 
The tuner uses cost model from each family to estimate the latency of each candidate, and selects the candidates with higher improving potential.
The selected candidates are then sent to real hardware (\eg, GPUs) for latency evaluation, and the results are returned to the subgraph family. The evaluation results are forwarded to the cost model inside the family, behaving as the training samples to help improve the accuracy of the cost model.
Since subgraphs within a family share similar characteristics, we can use the cost model to foresee other subgraphs, which replaces the costly code measurements on real hardware and saves the time budgets, and thus improves the search efficiency.
This procedure is executed iteratively, till the time budgets are used up.

\section{methodology}
\label{sec:methodology}
In this section, we describe the methods to explore the similarity of subgraphs and construct subgraph families. Then we  present the algorithms of FamilySeer to apply subgraph families to optimize the allocation of time budget so as to improve the tuning efficiency and quality.

\subsection{Identifying Similar Subgraphs}
\label{subsec:iss}
We find out that there are similarities between different subgraphs (described in Section~\ref{sec:motivation}).
To generate an accurate cost model for the similar subgraphs, we need to cluster the subgraphs and reorder into subgraph families. 
The similarity between subgraphs lies on the accuracy across the individual cost models of individual subgraphs. 
Our goal is to find several subgraph families so that the subgraphs can benefit from the family cost models with higher estimation accuracy. 
An intuitive approach is to apply the individual cost model of each subgraph to estimate program candidates of other subgraphs, and classify the subgraphs with impressive accuracy into a family, during tuning. This approach ensures that each subgraph can be placed on the most suitable subgraph family. 
However, it requires continuous estimations on real hardware and redundant training of cost models of all subgraphs, which introduces non-trivial overhead. 
Moreover, due to the limited number of training samples at the beginning, the cost models lack the training data and usually fails to perform reliable estimations. This may result in two dissimilar subgraphs being assigned together, which affects the quality of the search.
Therefore, FamilySeer adopts the approach of static analyzing, which classifies the subgraphs based on their attributes (\eg, operation sequence, input shape, \etc).
There is no overhead in the tuning process because the classification can be done ahead of tuning.

\begin{figure}[htbp]
	\centering
	\includegraphics[width=\linewidth]{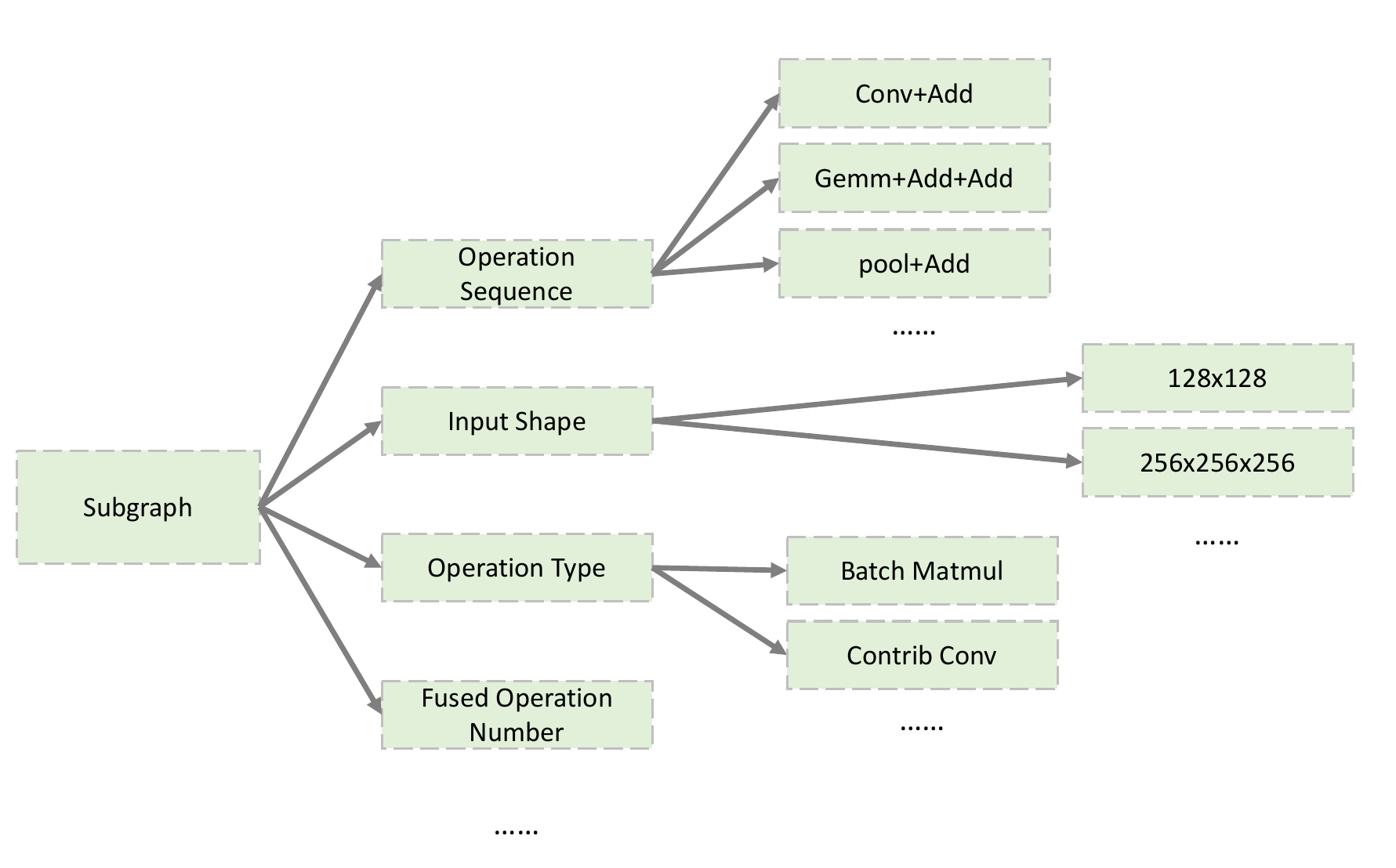}
	\caption{Attributes of the subgraphs.}
	\label{fig:Attributes}
\end{figure}

We try to analyze the attributes of these subgraphs and find out these subgraph families.
A subgraph contains a variety of different operators with different input shapes, which form different subgraphs based on a certain sequence. 
Most of the deep learning compiler generate many subgraph attributes to help to distinguish different subgraphs. 
As shown in Figure ~\ref{fig:Attributes}, the operator sequence of the subgraph is a description to the data flow of the subgraph, which is unrelated to the size and shape of the subgraph. 
The compilers tend to convert these sequences into strings using serialization algorithms (\eg, \textit{hash}) and using these serialized strings to distinguish different subgraphs. 
The compilers also record the core operators when performing operator fusion.  
Therefore, there are three algorithms of static analyzing based on the subgraph attributes: \textit{1)} by the number of operators, \textit{2)} by core operator, and \textit{3)} by the operator sequence.
We use these three algorithms to generate subgraph families and analyze the accuracy of each subgraph using its cost model inside the subgraph family. 
As shown in Table ~\ref{tab:method_analyzing}, using the algorithm by the core operation to generate subgraph families achieves the highest accuracy (99.4\%). Therefore we construct subgraph families by the core operation of all subgraphs.
Notably, Family still reserves an interface for static analyzing. If better algorithms are explored  in the future, we can quickly replace the current algorithm (by the core operation) with them.

\begin{table}[htbp]
    \centering
    \caption{Cost model accuracy under different algorithms of static analyzing.}
    \label{tab:method_analyzing}
    \begin{tabular}{lc}
        \hline
        Algorithm           & Accuracy       \\ \hline
        By operation number & 97.4\%          \\
        By core operation    & \textbf{99.4\%} \\
        By operation sequence    & 99.1\%          \\ \hline
        \end{tabular}
\end{table}

\subsection{Foresee tuning}

As shown in Figure ~\ref{fig:overview}, subgraphs inside the same subgraph family share the same cost model. We can use this cost model to \textbf{foresee} better candidates for each subgraph and generate optimal program candidates. 
The algorithm of foresee tuning is shown in Algorithm~\ref{alg:foresee}. 
When tuning a deep learning model $N$, users usually provide time budget $B$ (\ie, number of measurements on real hardware) or an expected latency of model inference.  
The foresee proportion $p$ determines the tuning opportunity that can be shared between bottleneck subgraphs and less-improved subgraphs inside the same subgraph family, which is described in detail later.
The foresee tuning algorithm takes $N$, $B$, and $p$ as the inputs. 

The operators of model $N$ is fused according the pre-defined fusion rules of TVM, and each fused operators are considered as a subgraph  (line~\ref{alg:line:construct_subgraph}).
Then the subgraph families are constructed by identifying the similar subgraphs (line~\ref{alg:line:construct_family}), as described in Section~\ref{subsec:iss}, and an individual cost model is initialized for each subgraph family (line~\ref{alg:line:init_model}).
The number of generated candidates of each bottleneck subgraph during the tuning iteration is defined in line~\ref{alg:line:g}, which is similar with the mini-batch size. This number is restricted to 64 at most, in order to increase the tuning iterations and avoid the over-fitting.

When the used budget $b$ (initialized to 0, line~\ref{alg:line:b}) is less than the time budget $B$, FamilySeer conducts two tuning steps, one for the bottleneck subgraph of all subgraphs and one for the bottleneck graph of the subgraph family. This procedure is conducted in an iterative manner until $b \ge B$.
Firstly, FamilySeer focuses on the all subgraphs, and calculates the improving potential of all subgraphs (line~\ref{alg:line:p}). Specifically, the subgraphs with higher latency have higher improving potential. 
Then the subgraphs are sorted by their potential in descending order (line~\ref{alg:line:sort}), so that the top subgraph (denoted as $s_{cur}$) is the bottleneck subgraph with the highest latency (line~\ref{alg:line:pop}). 
Then the subgraph family (line~\ref{alg:line:find_family}) of the bottleneck subgraph and the corresponding cost model (line~\ref{alg:line:find_model}) are figured out.
FamilySeer tunes this subgraph with the help of the family cost model (line~\ref{alg:line:tune}). 
The tuner inside $tune$ function generates program candidates for subgraph $s_{cur}$ and then use the family cost model $c_{cur}$ to estimate the latency of the candidates. 
The candidates with lower latency will be evaluated on the real hardware, and the number of candidates is restricted to $g$.
The evaluated candidates are used to update the family cost model and help to improve accuracy of the cost model for the next tuning iteration (line~\ref{alg:line:train}).

Then, FamilySeer focuses on the current subgraph family ($f$) and begins the foresee tuning. 
Subgraphs inside the same family share similar code structures and therefore their program candidates tend to be estimated accurately with the shared family cost model. 
If the family contains only one subgraph (\ie, $s_{cur}$), which mean the current subgraph has no similar subgraph, the foresee tuning is skipped.
Otherwise, FamilySeer restricts the scope to the subgraph family $f$ and repeats the core tuning steps from line~\ref{alg:line:core_bgn} to line~\ref{alg:line:core_end}.
Notably, the number of generated program candidates is set to $g \times p$ (line~\ref{alg:line:tune}). 
The adjustable foresee proportion $p$ determines the proportion of tuning round shared between the bottleneck subgraph and less-bottleneck subgraph inside the same subgraph family, and FamilySeer appends $g \times p$ program candidates to the subgraph family in the foresee tuning.

The value of $p$ should be greater than 0 but less than 1. A larger $f$ increases the time budget of tuning less-bottleneck subgraph, but may cause spending too much time and thus trivializing the bottleneck subgraphs.
A smaller $p$ has less effect on the entire tuning process, but it requires a highly accurate cost model to estimate the program candidates. 
Therefore we recommended to set a larger $p$ if users allows sufficient tuning time, and vice versa.
By default, FamilySeer sets $p$ to 0.25, which we think  has acceptable overhead on the tuning time while is enough to adapt to the family cost model. 

\begin{algorithm}[htbp]  
	\caption{Foresee tuning}
	\label{alg:foresee}
	\begin{algorithmic}[1]  
		\Require DL model $N$, time budgets $B$ 
		\Require $p \gets$  adjustable foresee proportion 
		\State {$b = 0$} \label{alg:line:b}
		\Comment {\# of used budgets}
		\State {subgraphs = construct\_subgraphs($N$)}  \label{alg:line:construct_subgraph}
		\State {families = construct\_family(subgraphs)} \label{alg:line:construct_family}
		\For {$f$ in families}
		\State {$C[f]$ = initialize\_cost\_model($f$)}
		\EndFor  \label{alg:line:init_model}
		\State {$g = \min(64, B/\mbox{\# of subgraphs})$} \label{alg:line:g}
		\Comment {generated candidates in each tuning iteration}
		\While{$b$ < $B$}
		\State {$scope$ = subgraphs}
		\Comment {consider all subgraphs}
		\For {$s$ in $scope$} \label{alg:line:core_bgn}
		\Comment {\textbf{core tuning steps}}
		\State {$P[s]$ =  calculate\_potential($s$)} \label{alg:line:p}
		\EndFor
		\State {sorted($scope$, key=<potential>, order=desc)} \label{alg:line:sort}
		\State {$s_{cur}$ = $scope$.pop()} \label{alg:line:pop}
		\State {$f$ = find\_family($s_{cur}$)} \label{alg:line:find_family}
		\Comment {get the family of $s_{cur}$}
		\State {$c_{cur}$ = $C[f]$} \label{alg:line:find_model}
		\Comment {get the cost model of $s_{cur}$}
		\State {$g$ = $g$ if $scope$ = all subgraphs, $g \times p$ if $scope$ = family}
		\State {$result$ += $tune$($s_{cur}$, $c_{cur}$, candidates=$g$)} \label{alg:line:tune}
		\State {$c_{cur}$ = train\_cost\_model($result$, $c_{cur}$)}  \label{alg:line:core_end} \label{alg:line:train}
		\Comment {\textbf{core steps end}}
		\State $b$ += $g$
		\Comment {update $b$}
		\If {len($f$) > 1}
		\Comment {\textbf{begin foreseeing}}
		\State {$scope$ = $f$}
		\Comment {consider current subgraph family}
		\State {\textbf{repeat} line~\ref{alg:line:core_bgn} $\to$ line~\ref{alg:line:core_end}}
		\Comment {foresee the family}
		\State $b$ += $g \times p$
		\Comment {update $b$}
		\EndIf

		\EndWhile
	\end{algorithmic}  
\end{algorithm}

\subsection{Multi-GPU Acceleration}
Modern servers are equipped with multiple GPUs, allowing multiple tasks to be deployed simultaneously. During the tuning process, the tuner generates several program candidates and evaluates them on real hardware. 
Even if we optimize the tuner with foresee tuning, evaluating the candidate still consumes most of the tuning time. 
The candidates are transformed programs of the subgraphs and can be considered as independent tasks, therefore we can evaluate them on GPUs in parallel. 
We use remote procedure call (RPC) service to utilize multiple GPUs in a single server. Each GPU with the server is registered as a RPC device. 
As the tuning begins, the tuner generates several candidates and let the compiler transform these candidates into transformation codes (\eg, CUDA codes). The RPC service allocates each code to the registered GPU and collect the its real latency. 
The RPC service manages the allocation frequency to avoid evaluating multiple tasks on a single GPU at the same time, which helps to avoid the latency interference across the tasks. 

We also optimize the training efficiency of the cost model. 
The tuner relies on the cost model to estimate candidates during each tuning iteration, and the training data for the cost model comes from the evaluating result of each candidate. 
We keep training the cost model during the tuning procedure in order to gain better accuracy on the subgraphs. 
Training the machine learning cost model (\eg, XGBoost) on CPU only takes less than a second even under thousands of evaluated candidates~\cite{mitchell2018xgboost}. 
However, most of the deep learning models has more than ten subgraphs, which generate more than ten thousands of the evaluated candidates. 
As the tuning continues, each cost model are trained with more evaluated candidates, which consumes dozens of seconds.

To minimize the training time of cost models, we move the cost model from CPU to GPU. This method will not affect the process of evaluating candidates since the cost model is trained after the candidates have been evaluated.

\section{Evaluation}
\label{sec:evaluation}
In this section, we are trying to answer the following question: \textit{1)} How does FamilySeer compare to Ansor improve search efficiency? \textit{2)} What is the actual performance curve during the tuning process? \textit{3)} Can FamilySeer further improve the search quality?

\subsection{Experimental Setup}
We choose several Deep learning models and benchmark them in CPU and GPU platforms (shown in Table ~\ref{tab:platform}). The CPU platform is a dual-socket CPU node, each socket is a 10-core Intel Xeon Sliver 4210 with hyper-threading enabled. Each CPU applies 256 GB of RAM in four memory channels. The GPU platform is a dual-socket GPU cluster with two NVIDIA V100 32G, each socket is a 14-core Intel Xeon E5 2680 V4 with hyper-threading disabled. We compare the search quality and search efficiency against the state-of-the-art auto-tuning framework: Ansor (Commit: \textit{$64c1b79$}). The chosen models are shown in Table ~\ref{tab:model}. ResNet and Mobilenet represent traditional image classification. BERT, RoBERTa ,and GPT2 represents language translation. Vision Transformer (ViT) represents image classification based on transformer. The batch size is set to 1 since most inferences use one batch.

We let Ansor run the searching on its recommended time budgets. Ansor suggests 900 for each subgraph on GPU while 800 on CPU. For example, the ResNet50\_v1 has 28 subgraphs. The recommended time budgets is set to 25200. We set the beginning of the tuning where each subgraph has been tuned once so the cost model can be built.

\begin{table}[]
	\caption{The DL models used for evaluation.}
	\centering
	\label{tab:model}
	\begin{tabular}{clc}
	\hline
	\multicolumn{1}{l}{Model}                  & \multicolumn{1}{c}{Model}               & Task                                                                            \\ \hline
	\multicolumn{1}{c|}{\multirow{4}{*}{CNN}}         & \multicolumn{1}{l|}{ResNet50\_v1~\cite{he2016deep}}        & \multirow{5}{*}{\begin{tabular}[c]{@{}c@{}}Image\\ Classfication\end{tabular}}  \\
	\multicolumn{1}{c|}{}                             & \multicolumn{1}{l|}{ResNet152\_v2~\cite{he2016identity}}      &                                                                                 \\
	\multicolumn{1}{c|}{}                             & \multicolumn{1}{l|}{Mobilenet~\cite{howard2017mobilenets}}      &                                                                                 \\
	\multicolumn{1}{c|}{}                             & \multicolumn{1}{l|}{Mobilenetv2 ~\cite{sandler2018mobilenetv2}}    &                                                                                 \\ \cline{1-1}
	\multicolumn{1}{c|}{\multirow{4}{*}{Transformer}} & \multicolumn{1}{l|}{ViT-Huge~\cite{dosovitskiy2020image}} &                                                                                 \\ \cline{3-3} 
	\multicolumn{1}{c|}{}                             & \multicolumn{1}{l|}{BERT-Large~\cite{devlin2018bert}}               & \multirow{3}{*}{\begin{tabular}[c]{@{}c@{}}Language\\ Translation\end{tabular}} \\
	\multicolumn{1}{c|}{}                             & \multicolumn{1}{l|}{RoBERTa-Large~\cite{liu2019roberta}}            &                                                                                 \\
	\multicolumn{1}{c|}{}                             & \multicolumn{1}{l|}{GPT2-Small~\cite{radford2019language}}               &                                                                                 \\ \hline
	\end{tabular}
	\end{table}

\begin{table}[]
	\caption{The hardware platforms used for evaluation.}  
	\centering  
	\label{tab:platform}
	\begin{tabular}{cll}
	\hline
	\multicolumn{1}{l}{}        & \multicolumn{1}{c|}{CPU Platform}      & \multicolumn{1}{c}{GPU Platform} \\ \hline
	\multicolumn{1}{c|}{CPU}    & \multicolumn{1}{l|}{Intel Sliver 4210} & Intel E5 2680 V4                 \\
	\multicolumn{1}{c|}{GPU}    & \multicolumn{1}{l|}{N/A}               & NVIDIA V100 32G                  \\
	\multicolumn{1}{c|}{RAM}    & \multicolumn{1}{l|}{DDR4 2666 512G}    & DDR4 2400 384G                   \\ \cline{2-3} 
	\multicolumn{1}{c|}{System} & \multicolumn{2}{c}{Ubuntu20.04}                                           \\
	\multicolumn{1}{c|}{GCC}    & \multicolumn{2}{c}{9.3}                                                   \\ \cline{2-3} 
	\multicolumn{1}{l|}{CUDA}   & \multicolumn{1}{l|}{N/A}               & 11.2                             \\
	\multicolumn{1}{l|}{Driver} & \multicolumn{1}{l|}{N/A}               & 460.80                           \\ \hline
	\end{tabular}
	\end{table}

\subsection{Search Efficiency Improvement}
\label{subsec:sei}
\begin{figure*}
	\centering
	\subfigure[On Xeon Sliver 4210]
	{
		\centering
		\includegraphics[width=0.98\linewidth]{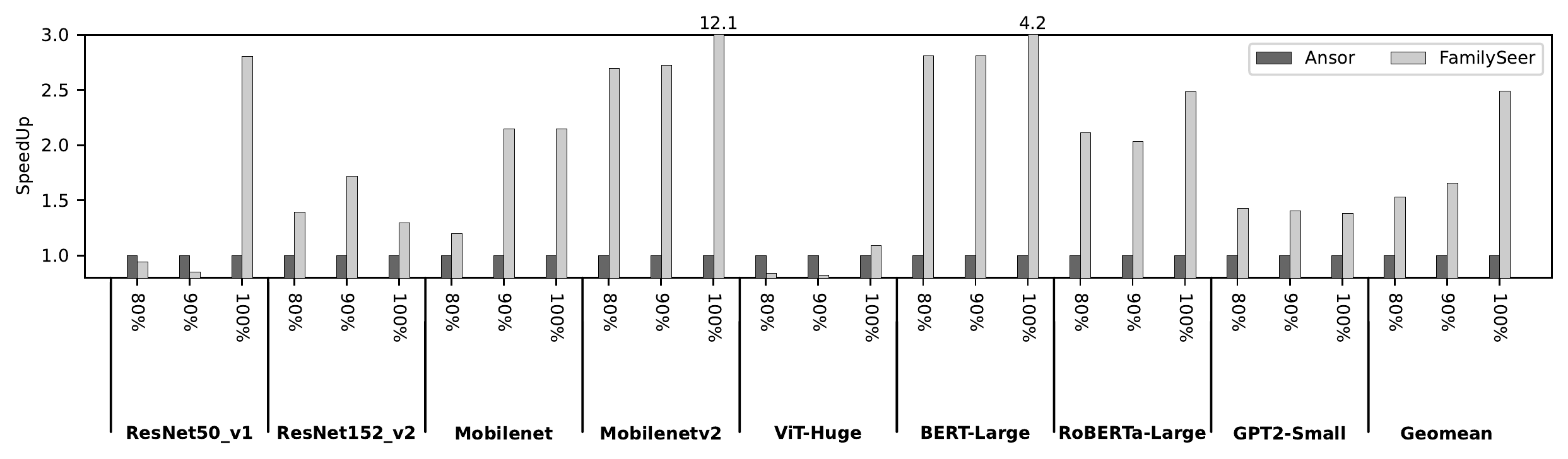}
		\label{fig:speedup:cpu}
	}
	\subfigure[On NVIDIA V100]
	{
		\centering
		\includegraphics[width=0.98\linewidth]{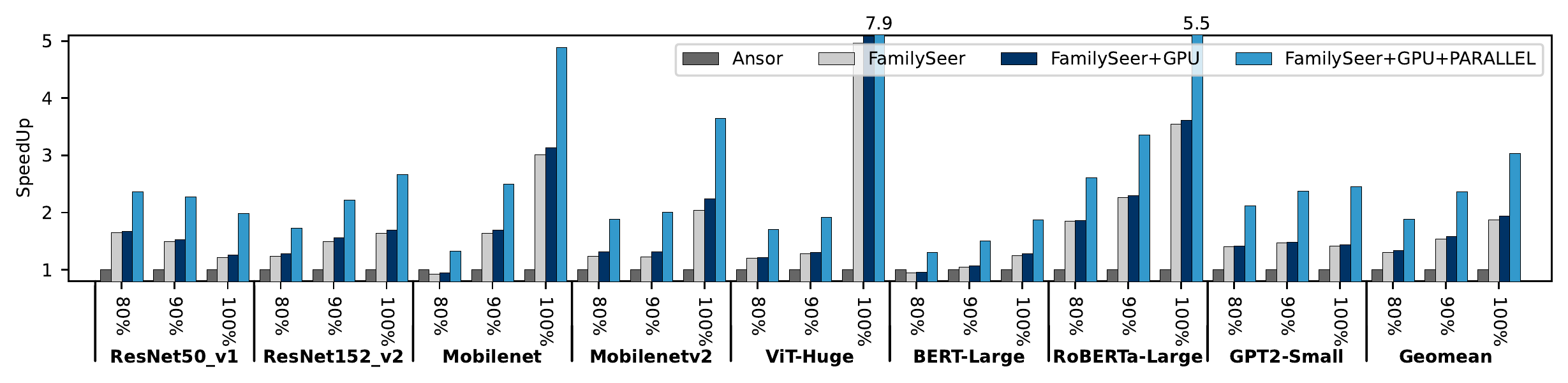}
		\label{fig:speedup:gpu}
	}
	\caption{Speedup comparison between Ansor and FamilySeer. We compare the used time of FamilySeer and Ansor when reaching Ansor's 80\%, 90\% and 100\% performance of the inference time.}
	\label{fig:speedup}
\end{figure*}

The search efficiency can be described as the tuning time to an optimal end-to-end inference time. The higher the search efficiency is, the lower end-to-end inference time is reached. To demonstrate the convergence of end-to-end performance, we choose three end-to-end inference times from Ansor when getting 80\%, 90\%, and 100\% of the inference time as our baseline. The results are shown in Figure ~\ref{fig:speedup}.

Figure ~\ref{fig:speedup:cpu} demonstrates the tuning time speedup on CPU. The search efficiency of Ansor (black bars) is normalized to 1. FamilySeer (gray bars) reduces search time by up to 12.1$\times$ while still reaching the same search quality. Overall, FamilySeer has an average of {1.53$\times$, 1.66$\times$, 2.49$\times$} speedup on 80\%, 90\% and 100\% of performance, respectively. 
Model likes Mobilenetv2 is partitioned into several Conv2D subgraphs with relatively close execution time, Softmax subgraph, and Pooling subgraph. FamilySeer forms these subgraphs into several subgraph families to provide pure training data for the cost model inside the families. 
The highly accurate cost model helps improve search efficiency. We can also find that the search efficiency of models (Mobilenet, Bert-Large, RoBERTa-Large, and GPT2-Small) has better improvement as the search process continue. This is because these models tend to have many similar subgraphs and can easily benefit from FamilySeer. 
Model likes ResNet50\_v1 has varied kinds of Conv2D such as Contrib Conv2D. The diversity of the subgraphs requires more candidate evaluation before the cost model in the family can search for better candidates, which results to a slight performance lag (0.85$\times$ at 90\%) before converging. 
As the search process continues, the training data enrich and performance regain. 
ViT-Huge also has a slight performance lag (0.82$\times$ at 90\%) before converging. This is because evaluating subgraphs of ViT-Huge on CPU requires more time than other models. When the cost model has low accuracy with limited evaluated candidates, many non-improving candidates are measured, resulting in a longer time to reach the same end-to-end performance.

Figure ~\ref{fig:speedup:gpu} shows the tuning time speedup on GPU. Because the GPU platform has 2 GPUs, we can measure the candidates parallelly and speed up the training of the cost model. The dark blue bar chart represents FamilySeer accumulated with the speedup of training cost model on GPU. The blue bar chart uses all the techniques. It shows that FamilySeer can reduce up to 7.9$\times$ search time and has an average of 1.89$\times$, 2.36$\times$, 3.04$\times$ speedup on 80\%, 90\% and 100\% of performance, respectively. FamilySeer has demonstrated a stable improvement compared to Ansor in most models except ResNet50\_v1. We find that Both methods have similar peak performance, but FamilySeer reaches the peak performance of the Resnet50\_v1 much earlier than Ansor. Therefore, FamilySeer enlarges the gap between Ansor at the beginning. But as the tuning continues, the gap shrinks until Ansor runs out of its time budget.

When moving the cost model from CPU to GPU, we can also find extra performance improvement from 1.01$\times$ to 1.09$\times$ because training cost model on GPU is much faster than on CPU when the training data is extensive. The improvement becomes more significant as long as the model has more subgraphs. For example, model likes MobileNetv2 has more than 30 subgraphs. Thus it has more than 30000 candidates as training data, resulting in a speedup of 1.09$\times$ and thousands of seconds being saved. Model likes RoBERTa-Large have less than ten subgraphs. The speedup is 1.01$\times$ and hundreds of seconds have been saved. Training cost model on GPU can also benefit from a longer time budget, meaning more candidates are measured and treated as training data.

The speedup of using parallel GPU is from 1.56$\times$ to 1.79$\times$ compared to measuring candidates sequentially. Paralleling measurement on GPU can benefit those subgraphs which have longer execution time. For example, most of the execution time of the subgraphs in GPT2-Small are more than dozens of milliseconds and therefore has a speedup of 1.72$\times$. Having more subgraphs and evaluating candidates also benefit from parallel measurement such as MobileNetv2 (1.79$\times$). The speedup of RoBERTa-Large is 1.56$\times$, which is the lowest among all the other models. This is because RoBERTa-Large has less than ten subgraphs and the execution time of each subgraph is less than a millisecond.

\subsection{Search Quality Improvement}
Although both Ansor and FamilySeer search on the same search space, we explore the search space differently. We give both Ansor and FamilySeer sufficient time budget and compare end-to-end performance.

Table ~\ref{tab:e2e_performance_cpu} and ~\ref{tab:e2e_performance_gpu} show end-to-end performance on CPU and GPU respectively. Models like Mobilenetv2 on GPU has shown an improvement of 8\%. Other models have also shown an improvement of 1-2\%. 
The result on CPU shows better performance by up to 1.17$\times$ on Mobilenet. 
Note that as the searching continues, the peak performance of the subgraphs is explored, and it makes even 1\% of the improvement difficult. Although we still search these candidates on the same search space, the difference between Ansor and FamilySeer is the time converges to the same performance. Ansor may still reach the same performance as long as it has been given more searching budget or explore the entire search space.

\subsection{Turing Performance Curve}

\begin{figure}[htbp]
	\centering
	\includegraphics[scale=0.7]{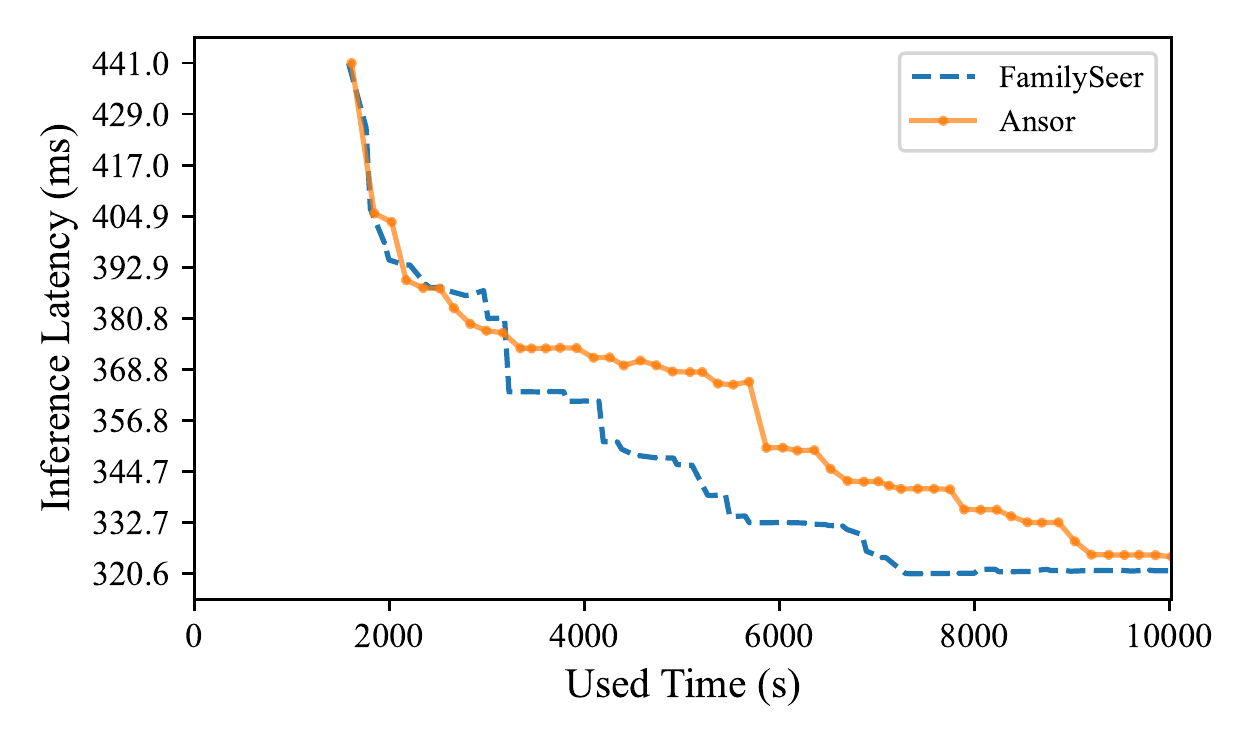}
	\caption{End-to-end latency curve of Ansor and FamilySeer when tuning GPT2-Small  on NVIDIA V100.}
	\label{fig:curve}
\end{figure}

We analyze the tuning curve of Ansor and FamilySeer. We evaluate and report the end-to-end latency change each time better candidates have been generated to show the actual end-to-end inference time during the tuning process. Figure ~\ref{fig:curve} shows the tuning curve of GPT2-Small on GPU. When the search begins, FamilySeer is almost the same compared to Ansor. As the tuning continues, each subgraph family gains sufficient training data and therefore, better prediction accuracy. We can see many nearly vertical curves on FamilySeer. This is because FamilySeer utilizes foresee tuning and subgraphs inside the same families can get better candidates under a smaller time budget. Eventually, both Ansor and FamilySeer reach the bottleneck of our tuning performance, and we can have better performance compared to Ansor.

\begin{table}[]
	\caption{End-to-end performance comparison on CPU.}  
	\centering
	\label{tab:e2e_performance_cpu}
	\begin{tabular}{llll}
		\hline
											 & \multicolumn{1}{c}{Ansor} & \multicolumn{1}{c}{FamilySeer} & \multicolumn{1}{c}{Speedup} \\ \hline
		\multicolumn{1}{l|}{ResNet50\_v1}     & 12.99 ms                  & 12.84 ms     & 1.012x                  \\ \hline
		\multicolumn{1}{l|}{ResNet152\_v2}    & 36.17 ms                  & 36.08 ms     & 1.002x                  \\ \hline
		\multicolumn{1}{l|}{Mobilenet}   & 1.132 ms                  & 0.989 ms     & 1.145x		\\ \hline
		\multicolumn{1}{l|}{Mobilenetv2} & 1.79 ms                   & 1.53 ms      & 1.170x                  \\ \hline
		\multicolumn{1}{l|}{ViT-Huge}        & 432.451 ms		& 422.504 ms		& 1.024x                 \\ \hline
		\multicolumn{1}{l|}{BERT-Large}      & 149.1 ms                  & 146 ms        & 1.021x                 \\ \hline
		\multicolumn{1}{l|}{RoBERTa-Large}   & 146.4 ms                  & 141.4 ms      & 1.035x                 \\ \hline
		\multicolumn{1}{l|}{GPT2-Small}      & 5049 ms                   & 4945 ms       & 1.021x                 \\ \hline
	\end{tabular}
\end{table}

\begin{table}[]
	\caption{End-to-end performance comparison on GPU.}  
	\centering  
	\label{tab:e2e_performance_gpu}
	
		\begin{tabular}{llll}
		\hline
											 & \multicolumn{1}{c}{Ansor} & \multicolumn{1}{c}{FamilySeer} & \multicolumn{1}{c}{Speedup} \\ \hline
		\multicolumn{1}{l|}{ResNet50\_v1}     & 1.8 ms                    & 1.78 ms      & 1.011x                  \\ \hline
		\multicolumn{1}{l|}{ResNet152\_v2}    & 5.42 ms                   & 5.24 ms       & 1.034x                 \\ \hline
		\multicolumn{1}{l|}{Mobilenet}   & 0.244 ms                  & 0.241 ms     & 1.011x                   \\ \hline
		\multicolumn{1}{l|}{Mobilenetv2} & 0.37 ms                   & 0.34 ms      & 1.087x                  \\ \hline
		\multicolumn{1}{l|}{ViT-Huge}        & 50.76 ms                  & 50.56 ms      & 1.004x                 \\ \hline
		\multicolumn{1}{l|}{BERT-Large}      & 17.99 ms                  & 17.85 ms       & 1.008x                \\ \hline
		\multicolumn{1}{l|}{RoBERTa-Large}   & 17.67 ms                  & 16.98 ms      & 1.041x                \\ \hline
		\multicolumn{1}{l|}{GPT2-Small}      & 324.7 ms                  & 320.7 ms      & 1.013x                 \\ \hline	
	\end{tabular}
\end{table}

\section{Related Work}

Deep learning compilers take deep learning Model as input and optimize the execution of deep learning model. The compilers describe computation and scheduling using their own intermediate representation. There are many Well known deep learning compilers such as XLA~\cite{abadi2017computational}, nGraph~\cite{cyphers2018intel}, TVM~\cite{chen2018tvm}, TACO, Tensor Comprehensions~\cite{vasilache2018tensor}, Halide~\cite{ragan2013halide}, Tiramisu~\cite{baghdadi2019tiramisu} and Glow~\cite{rotem2018glow}.

Many techniques have been introduced to optimize deep learning model on Deep learning compiler. AutoTVM~\cite{chen2018learning} is a templated-guided search framework using handwritten template to optimize computation definition. The template defines the structure of the tensor expression and provide several tunable parameters. The compiler search and match these expressions with the template and tuning for optimal parameter. The size of the search space is defined by the template itself. 

Ansor~\cite{zheng2020ansor} is a search-based framework using cost model to guide and optimize deep learning model. The compiler partition model into multiple subgraphs and framework generate many transformation codes according to the optimization rules. These transformation codes are estimated by the cost model to find the possible better code. 

The large search space defined by the optimization rules mean the accuracy of the cost model will affect the efficiency of the search-based framework directly. Thus, many works focus on providing highly accurate cost model like Tenset~\cite{zheng2021tenset} use pretrained cost model. Some works optimize the cost model itself~\cite{steiner2021value,kaufman2021learned,baghdadi2021deep}. Some research focus on subgraph optimization such as fusing operation and \etc{~\cite{fang2020optimizing,jia2019optimizing,jia2019taso,wang2021pet}}.

\label{sec:related}

\section{Conclusion}
\label{sec:conclusion}
We propose FamilySeer, a new auto-tuning framework for deep learning compilers to exploit the similarity of subgraphs and generate optimal code transformations. We evaluate the possibility of forming a similar subgraph into a subgraph family to improve the search quality and efficiency for the state-of-art auto-tuning frameworks. Furthermore, we also utilize the advantage of the subgraph families to accelerate the converge to optimal code. FamilySeer outperforms the existing searching framework by up to 3.04$\times$ on the search efficiency and further explores the search quality by up to 1.17$\times$. We hope that FamilySeer can help improve the search efficiency of the auto-tuning in deep learning compiler to become more efficient.

%

%
\bibliographystyle{ACM-Reference-Format}
\bibliography{reference.bib}

%
\end{document}